\documentclass[runningheads]{llncs}

\usepackage{graphicx}
\usepackage{hyperref}
\usepackage{multirow}

\begin{document}

\title{Towards Financial Sentiment Analysis in a South African Landscape}
%
%
\author{Michelle Terblanche\orcidID{0000-0002-1460-3596} \and
Vukosi Marivate\orcidID{0000-0002-6731-6267}}
\institute{Department of Computer Science\\ University of Pretoria\\ South Africa\\
\email{michelle.terblanche@gmail.com, vukosi.marivate@cs.up.ac.za}}
%
\maketitle              
\begin{abstract}
Sentiment analysis as a sub-field of natural language processing has received increased attention in the past decade enabling organisations to more effectively manage their reputation through online media monitoring. Many drivers impact reputation, however, this thesis focuses only the aspect of financial performance and explores the gap with regards to financial sentiment analysis in a South African context.  Results showed that pre-trained sentiment analysers are least effective for this task and that traditional lexicon-based and machine learning approaches are best suited to predict financial sentiment of news articles. The evaluated methods produced accuracies of 84\%-94\%. The predicted sentiments correlated quite well with share price and highlighted the potential use of sentiment as an indicator of financial performance. A main contribution of the study was updating an existing sentiment dictionary for financial sentiment analysis. Model generalisation was less acceptable due to the limited amount of training data used. Future work includes expanding the data set to improve general usability and contribute to an open-source financial sentiment analyser for South African data.

\end{abstract}
\section{Introduction}\label{intro}
Big corporate organisations produce vast amounts of textual information in the form of official financial and non-financial reports, media releases and trading statements. 
The communication strategy, and hence perception, of an organisation directly impacts it's reputation. One of the industry accepted measures of reputation, {\it the RepTrak Score}\footnote{https://www.reptrak.com/reputation-intelligence/what-is-it/}, takes into account seven drivers of reputation: products and services, innovation, workplace, citizenship, governance, leadership and performance. The latter is a measure of the financial health of an organisation.


In the past, historic accounting information formed the basis for financial performance prediction, evolving from statistical models to more sophisticated machine learning models \cite{ref:Hajek2014}. Subsequent research ventured into the field of qualitative measures such as textual analysis to predict performance \cite{ref:loughran2011liability}. More recent research shows that there is promise in correlating sentiment with financial performance in order to make future predictions \cite{ref:Hajek2014,ref:Joshi2016,ref:loughran2011liability,ref:Mudinas2019}.   


The justification for this study is rooted firstly in the evidence that there is a financial value linked to the reputation of an organisation \cite{ref:lei2011financial,ref:Vig2017}. An improvement in reputation can have in the order of a 6\% improvement in the company bottom-line \cite{ref:lei2011financial}. As a result, reputation risk should form a key component of overall corporate strategy \cite{ref:Vig2017}. 


A further motivation for this study stems from identifying a gap in the South African context with regards to financial sentiment analysis using natural language processing (NLP) techniques. Even though many sentiment analysers are freely available, these models were developed within a given context and relevant to a specific domain and geographical region.

Based on the identified problem and motivation, the following research questions were identified:  \textit{What NLP techniques are required to successfully determine the sentiment of financial communication in a South African context?}; \textit{Is there a correlation between the sentiment of financial news and company performance as indicated by share price?}; \textit{How effectively can a narrower sentiment prediction model be applied to a broader scope of finance-related information?}  The study only focuses on formal communication channels in the form of online news articles, specifically excluding social media.

\section{Sentiment Analysis and Opinion Mining}\label{sentgen}
The terms sentiment analysis and opinion mining are often used interchangeably. The first mention of public opinion analysis dates back to post-World War II and has been one of the fastest developing areas in the last decade. It involves using natural language processing (NLP) techniques to extract and classify subjective information expressed through opinions or through detecting the intended attitude \cite{ref:Mantyla2018,ref:Saberi2017}. It has been one of the fastest developing areas in the last decade, growing from simple online product reviews to analysing the sentiment from various online platforms such as social media and extending the application to predicting stock markets, tracking polls during elections and disaster management \cite{ref:Mantyla2018}. Research highlighted the following three categories of sentiment analysis:
\paragraph{\textbf{Open-source pre-trained sentiment analysers}}
	The \textbf{\textit{TextBlob}} library in \textbf{Python} is a simple rule-based sentiment analyser that provides the average sentiment (excluding neutral words) of a text string\footnote{https://textblob.readthedocs.io/en/dev/}. 
	\textbf{\textit{VADER}}\footnote{Valence Aware Dictionary for sEntiment Reasoning} is another rule-based sentiment analyser specifically trained on social media texts and generalises quite well across contexts/domains compared with other sentiment analysers \cite{ref:Hutto2014}. This analyser outperforms \textbf{\textit{TextBlob}} when predicting sentiment on social media texts \cite{ref:Kumaresh2019}. Some of the main reasons are that it takes into account emoticons, capitalization, slang and exclamation marks.
\paragraph{\textbf{A simple dictionary-based approach}}
	This method typically uses a dictionary of words/phrases either manually created or automatically generated.

\paragraph{\textbf{Custom-built predictive models using machine learning}}
	The main techniques generally used involve either 1) traditional models or 2) deep learning models \cite{ref:Dang2020}. These are supervised machine learning models and required data sets to be labeled. The traditional models are typically Naive Bayes, logistic regression and support vector machines.\\

\noindent Some of the main challenges in sentiment analysis are language dependency, domain specificity, nature of the topic, negation and the availability of labeled training data \cite{ref:Hussein2018,ref:Saberi2017}. A further challenge in opinion mining from user generated content is to acknowledge the importance of text pre-processing to improve the quality and usability thereof \cite{petz2015reprint}. 

\section{Financial Sentiment Analysis}

\subsection{Exploiting Typical Financial Headline Structure} \label{hinge}
\begin{figure*}[h!]
	\centering
	\includegraphics[width=2.5in]{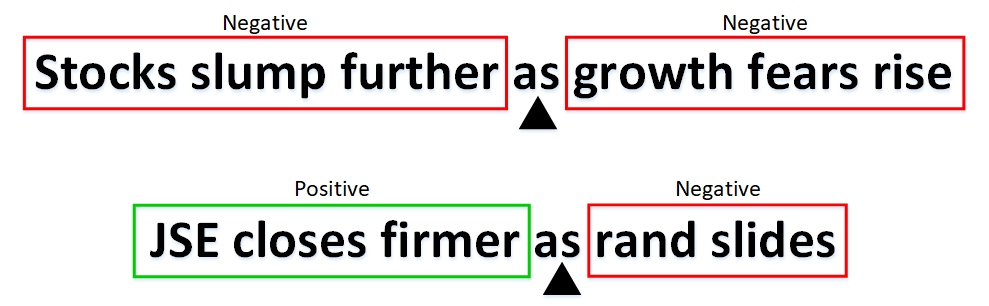}
	\caption{Hinge concept of financial headlines.}
	\label{fig:hinge}
\end{figure*}

A potential way to determine the sentiment of a financial title was explored by introducing the concept that $\pm$ 30\% of such titles follow a hinge structure \cite{ref:Zimmer}. The investigation suggested that the hinge, which is typically a word such as \textbf{\textit{as, amid, after}}, splits the sentence into two parts, both parts carrying the same sentiment. If one therefore determines the sentiment of the first part of the sentence, the overall sentiment is inferred. In Figure~\ref{fig:hinge}, the top sentence aims to explain this notion. However, the second sentence shows an example where the part of the sentence following the hinge does not carry the same sentiment as the first part.

Furthermore, it was argued that the verbs hold the key as sentiment carrying words. It was identified, however, that using existing, labeled word lists may still fall short since these lists were created using very domain-specific pieces of text. For e.g. a word such as rise may be listed as positive based on prior usage, however, its use in a new application may indicate it to be negative.

\subsection{Existing Approaches for Financial Sentiment Analysis}

\subsubsection{Existing Popular Financial Sentiment Word Lists}
For a lexicon-based approach, a very popular domain-specific (i.e financial) dictionary is the \textit{Loughran-McDonald sentiment word lists} first created in 2009 \cite{ref:loughran2011liability}. The drive for developing these lists stemmed from the authors showing that a  more general dictionary, in this case the \textit{H4N} negative wordlist from the \textit{Harvard Psychological Dictionary}, misclassified the sentiment of financial words quite substantially. They found that $\sim$75\% of negative words in the aforementioned list are generally not negative in the financial domain. The sentiment categories are negative, positive, uncertainty, weak modal, strong modal, litigious and constraining.

\subsubsection{Predictive models using machine learning}\label{machine}
As part of the 11$^{th}$ workshop on Semantic Evaluation (SemEval-2017)\footnote{https://alt.qcri.org/semeval2017/}, one of the tasks was \textit{``Fine-Grained Sentiment Analysis on Financial Microblogs and News"}\footnote{https://alt.qcri.org/semeval2017/task5/} of which a sub-task was sentiment analysis on news statements and headlines. It was a regression problem and participants had to predict the sentiment in the range -1 to 1 (representing Negative to Positive). The training data provided was annotated in this same range. Table \ref{tab:semeval} gives a summary of the results and methods for four of the submissions.

\begin{table}[h!]
	\caption{Summary of the performance of the various annotation methods.}
	\centering{
		\begin{tabular}[h]{|l|c|c|}
			\hline
			\bfseries Ranking & \bfseries Score$^{\mathrm{1}}$ & \bfseries Modelling Approach\\
			\hline
			1 & 0.745 & 1D convolutional neural network \\
			&&(using word embeddings from GloVe) \cite{ref:Mansar2017}\\
			\hline
			4 & 0.732&Bidirectional Long Short-Term Memory\\
			&& (with early stopping) \cite{ref:Moore2017}\\
			&& *Also looked at support vector regression\\
			\hline
			5 & 0.711&Ensemble using support vector regression\\
			&& (and gradient boosting regression) \cite{ref:Jiang2017}\\
			\hline
			8 & 0.695&Support vector regression\\
			&& (with word embeddings and lexicon features) \cite{ref:Kumar2017}\\
			\hline
			\multicolumn{3}{l}{$^{\mathrm{1}}$ Weighted cosine similarity score}\\
		\end{tabular}
	}
	\label{tab:semeval}
\end{table}

The models used to address the sentiment analysis task range from traditional machine learning to deep learning models with only a $\pm$5\% improvement from the latter (Table \ref{tab:semeval}). These results indicate that traditional machine learning models can be used quite successfully for this task to set a baseline for further evaluation. The prediction is still far better than random chance of 50\%. An important observation was the need for domain-specific sentiment lexicons. The challenge provided teams with labeled training data. However, a large number of supervised learning activities start with unlabeled data and a substantial amount of time and effort is required to properly annotate data sets

\subsection{Related Work on Financial Sentiment Analysis in the South African Context}

Research produced a limited amount of South Africa-related scientific papers on sentiment analysis as a sub-field of natural language processing. Even fewer published results were available on specifically financial sentiment analysis in a South African context.

In 2018, a study on using sentiment analysis to determine alternative indices for tracking consumer confidence (as opposed to making use of surveys) showed high correlation with the traditional consumer confidence indices  \cite{ref:Odendaal2018}. These indices are used to better understand current economic conditions as well as to predict future economic activity.



Another study, although not necessarily financial sentiment analysis \textit{per se}, was on measuring the online sentiment of the major banks in South Africa \cite{ref:Lappeman2020}. The data source for this analysis was social media only. Machine learning models were used for both detecting topics and analysing the sentiment of user-generated comments relating to those topics. The main contribution the authors made was to highlight the importance of human validation as part of the process to increase accuracy and precision \cite{ref:Lappeman2020}.


Based on the available research, it is deducted that a gap exists for researchers and academics to expand and improve sentiment analysis of online media through natural language processing, especially in the financial domain, in order to increase the knowledge base and pool of technical solutions in the context of South Africa.

\section{Sentiment Correlation with Financial Performance}




A statistical approach to understanding whether stock market prices follow a trend with the sentiment from news articles relating to the stock/company showed promising results \cite{ref:Chowdhury2014}. The method was tested on $\sim$15 different companies. The study only considered a dictionary-based approach to calculate degrees of positivity, negativity and neutrality. The results showed at 67\% correlation between sentiment and share price \cite{ref:Chowdhury2014}.


A second paper on predicting market trends using sentiment analysis included a broader context through more diverse data \cite{ref:Mudinas2019}. The authors evaluated a predictive model using sentiment attitudes (i.e. Positive and Negative), sentiment emotions (such as joy, anger) as well as common technical drivers of share price.  Granger-causality found that only sentiment emotions could potentially be useful indicators \cite{ref:Mudinas2019}.


The findings highlight the complexity of share price prediction and the fact that it is determined by a number of factors, of which sentiment could potentially add value. The authors highlighted the need to better understand which stocks are impacted by sentiment to determine to applicability of this proposed method \cite{ref:Mudinas2019}.



\section{Method} \label{method}

\begin{figure*}[h!]
	\centering
	\includegraphics[width=4.85in]{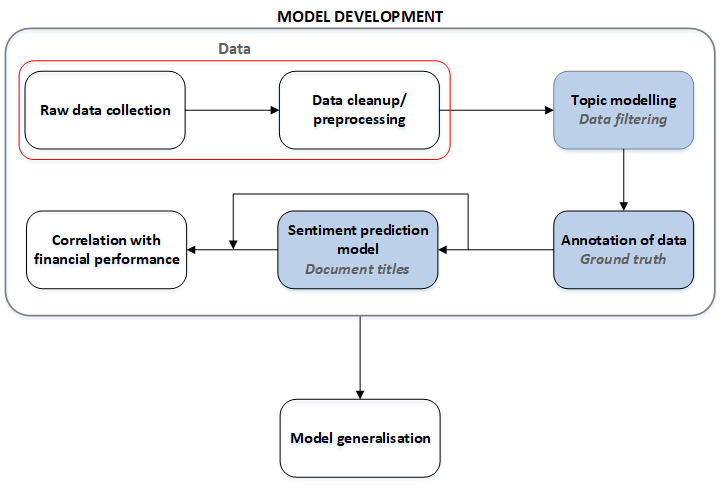}
	\caption{Process flow for addressing the research questions and objectives.}
	\label{fig:flow}
\end{figure*}

A model development pipeline was designed to answer the research questions and achieve the set objectives. This process flow is given in Figure~\ref{fig:flow}.

The following sections give a brief overview of the steps in the model development pipeline. 

During the first phase, \textbf{\textit{Data}}, various sources of publicly available textual information was identified and collected. Relevant data from these sources were extracted, cleaned and consolidated. The data sources used for model development are listed in Section \ref{data}.

\subsection{Topic Modelling for Data Filtering}

Topic modelling was used to filter the data to specifically extract financial-related documents. Term frequency–inverse document frequency (TF-IDF) was used to determine the word vectors (based on full article content) as input to the topic model. Non-Negative Matrix Factorization (NMF) was used.


\subsection{Annotation of Data}

Labeling of the data set, comprising only of the financial documents (extracted using topic modelling), was done on document headlines only using four independent annotators. Due to the small size of the data set, the full set of financial documents were labeled. Label options were Positive, Negative and Neutral. The majority label was used as the ground truth sentiment for the relevant documents.

The inter-annotator agreement was calculated using the \textbf{\textit{AnnotationTask class}}\footnote{https://www.nltk.org/\_modules/nltk/metrics/agreement.html} from NLTK in \textbf{Python}. Fleiss' Kappa was used as the statistical measure of inter-rater reliability \cite{ref:landis1977measurement}.

\subsection{Sentiment Prediction Model} \label{method}

Three sentiment prediction methods were evaluated, compared and the most robust prediction model implemented as the final annotation model.

\subsubsection{Existing rule-based Approaches: TextBlob and Vader}
Both \textbf{\textit{TextBlob}} and \textbf{\textit{VADER}} were used to calculate the sentiment of document headlines to understand the usability of pre-trained sentiment analysers on text from South African finance-related articles. Both models predict sentiment as value between -1 and 1 and it was assumed that predicted values between -0.05 and 0.05 are \textbf{\textit{Neutral}}.

\subsubsection{A simple lexicon-based approach}
For this analysis, the existing {\it Loughran and McDonald Sentiment Word Lists} were used as basis \cite{ref:loughran2011liability}.  These lists were developed to overcome the fact that more general dictionaries often misclassify financial texts, especially words perceived as negative in a day-to-day context. The following three iterations were performed:
\begin{enumerate}
	\item Experiment 1\\
	Base dictionary as updated in 2018\footnote{https://sraf.nd.edu/textual-analysis/resources/}.
	\item Experiment 2\\
	Base dictionary (Loughran and McDonald Sentiment Word Lists) with added synonyms (using NLTK's Wordnet Interface\footnote{https://www.nltk.org/howto/wordnet.html}). These synonyms are given the same sentiment.
	\item Experiment 3
	\begin{itemize}
		\item Base dictionary (as for experiment 2) but without the addition of synonyms for words in the "modal" lists. 
		\item Manual addition and deletion of words based on the evaluation of a sample of sentiment predictions from this update. 
	\end{itemize}
\end{enumerate}

The concept of a hinge structure (Section \ref{hinge}) using the words \textbf{\textit{as, but, amid, after, ahead, while}} and \textbf{\textit{despite}} was used. Where  hinge words were not present in headings, a \textbf{\textit{``comma''}} was used as a hinge or alternatively the full title was used. Individual words in article headings were lemmatized using multiple lemmas i.e. adjectives, verbs and nouns to ensure maximum chance of matching words in the developed dictionary.

Sentiment was assigned based on the first part of the title (where a hinge was present) alternatively the full sentence was used. This method assumes the sentiment is dictated by the first part, which could be slightly contradictory to the initial hinge structure proposal (Section \ref{hinge}).

In the case where multiple sentiment-carrying words are present, the first occurring \textbf{\textit{Positive}} or \textbf{\textit{Negative}} word was used as the sentiment of the headline (other sentiments were excluded in this round of the evaluation). Where no sentiment-carrying words were present, the headline was labeled as 'Not detected'. This approach does not take into account context, however, this simple bag-of-words implementation to detect word sentiments was used for the baseline model development. 

\subsubsection{Feature-based Approach: XGBoost}

A binary classifier was developed, using a traditional machine learning approach, with \textbf{Python's} implementation of \textbf{\textit{XGBoost}} (Extreme Gradient Boosting)\footnote{https://xgboost.readthedocs.io/en/latest/python/index.html}. It is a boosting algorithm based on an ensemble of decision trees\footnote{https://www.datacamp.com/community/tutorials/xgboost-in-python}.

The following typical cleaning and pre-processing steps were performed: tokenized text into words, converted words to lower case, expanded contractions (e.g. replace can't with can not), removed English stopwords, removed punctuation and lemmatized the words using NLTK's WordNetLemmatizer. Input vectors to the model were then created for the processed document headlines using term frequency inverse document frequency (TF-IDF).

Even though more advanced machine learning models have been used for sentiment classification (Section \ref{machine}), it was decided to only evaluate a more traditional machine learning model. The main reason being that the focus of the study was to develop an annotation method in order to set a baseline after which improvements can be investigated.

\subsection{Correlation with Financial Performance}
The predicted sentiments (from document headlines) and company financial performance (as indicated by share price), over the same time period, were analysed to observe whether patterns can be recognised. Multiple sentiments on a given date were resolved by using the majority sentiment. 

Since share price prediction is a complex task and impacted by various factors, it was decided to only illustrate whether a directional correlation can be observed. For future work, a statistical correlation can be investigated and potentially include additional drivers known to impact a given stock price. 

\section{Data} \label{data}
The company identified for developing the financial sentiment prediction model is \textbf{\textit{Sasol}}\footnote{www.sasol.com}. 

A variety of data sources were considered for model development and the most relevant were non-official communication in the form of online news articles and Stock Exchange News Service reports, which are company announcements that can have an affect on market movement. These are provided by the Johannesburg Stock Exchange (JSE)\footnote{https://www.jse.co.za/services/market-data/market-announcements} and are publicly available\footnote{https://www.sharedata.co.za}.

The above-mentioned data was collected for the period April/May 2015 - April/May 2020. The final data set were made up of 7666 online news articles and 168 SENS reports.

\section{Model Development Results}

\subsection{Annotation of Data}

Table \ref{tab:groundt} gives the sentiment distribution for the financial data set based on the majority label from the annotators. The 'None' category was removed.

\begin{table}[h!]
	\caption{Summary of the sentiment categories of the annotated data.}
	\centering{
		\begin{tabular}[h]{|l|c|c|}
			\hline
			\textbf{Sentiment} & \textbf{Count} & \textbf{Percentage} \\
			\hline
			Positive & 249&  31\%\\
			Negative & 419 & 52\%\\
			Neutral & 141 & 17\%\\
			\hline
		\end{tabular}
	}
	\label{tab:groundt}
\end{table}

The financial document data set, after using topic modelling for filtering and removing 'None' labeled documents, consisted of 808 articles (only 33 i.e. 4\% were SENS reports).

\subsection{Sentiment Prediction Model}

\subsubsection{Rule-based Approaches: TextBlob and Vader}
The \textit{Loughran and McDonald Sentiment Word Lists} only consider sentiment-carrying words, therefore to compare the various approaches only documents with \textbf{\textit{Positive}} and \textbf{\textit{Negative}} ground truth sentiments were considered (a total of 668 articles). Table \ref{tab:textb_full} shows the confusion matrices for using TextBlob and Vader to predict sentiment on document headlines. The overall accuracies were 19\% and 51\% respectively.The high inaccuracies stem from the majority of headlines being predicted as \textbf{\textit{Neutral}} (i.e. in the range -0.05 to 0.05).

\begin{table}[h!]
	\caption{Summary of the results using TextBlob and VADER on article headlines.}
	\centering{
		\begin{tabular}[h]{c|ccc||ccc|}
			&\multicolumn{3}{c||}{\textbf{TextBlob}}&\multicolumn{3}{c|}{\textbf{VADER}}\\
			\hline
			&&\multicolumn{2}{c||}{\textbf{Predicted}} &&\multicolumn{2}{c|}{\textbf{Predicted}} \\
			&&Negative&Positive&&Negative&Positive\\
			\hline
			
			\multirow{2}{*}{\textbf{Actual}}&Negative&68 &47&Negative&243& 86\\
			
			&Positive& 19 & 60&Positive& 53 & 99\\
		\end{tabular}
	}
	\label{tab:textb_full}
\end{table}

The superior performance of \textbf{\textit{VADER}} as compared with \textbf{\textit{TextBlob}} is consistent with a previous study on their comparison (Section \ref{sentgen}) \cite{ref:Hutto2014}. Furthermore, since \textbf{\textit{VADER}} was trained on social media, the subpar performance on financial headlines is therefore not unexpected.

\subsubsection{Lexicon-based Approach} \label{lexicon}

In Experiment 1, the original word lists (containing 4140 words) were used as is to determine a sentiment based on key words according to hinge structure approach discussed in Section \ref{hinge} to observe the baseline accuracy. The method for assigning  the sentiment to the headline is as outlined in Section \ref{method}.

The goal of Experiment 2 was to update the word lists with synonyms (of the words in the existing lists) and determine whether it improves prediction accuracy. As part of this experiment, a short list of bi-grams were added based on manual observation where one word was ambiguous (Table \ref{tab:bigram}). The list is not exhaustive and is to indicate the impact of expanding the sentiment dictionary.

\begin{table}
	\caption{Bi-grams added to the sentiment dictionary.}
	\centering{
	\small\addtolength{\tabcolsep}{5pt}
		\begin{tabular}[t]{|c|c|}
			\hline
			\bfseries {Negative} & \bfseries {Positive}\\
			\hline	   
			
			record low  &new record \\  
			record lows  &record high \\   
			back foot  &record highs \\   
			price halves  &record production \\   
			& on track\\
			
			\hline
		\end{tabular}
	}
	\label{tab:bigram}
\end{table}

Thereafter in Experiment 3, random samples were evaluated to update the dictionary from Experiment 2.  It was noticed that some of the synonyms added resulted in incorrect predictions and had to be removed again. Also, the synonyms added in this experiment excluded those for the ``modal" word lists. Only 4 words were removed from the original dictionary: break, closed, closing and despite. The final dictionary contains 9743  words.

It is recommended, however, that a more robust method be developed to update the dictionary in future since this manual method does not necessarily capture all the required words and may also have redundant words. Furthermore, due to the small size of the data set, manual updates could be performed but will not feasible for large data sets.
 
Table \ref{tab:lexicon} gives the results for the 3 experiments and highlights the improvement based on the manual dictionary update. After updating the dictionary, the sentiment prediction accuracy improved by 47\% compared with the original word lists. 

\begin{table}[h!]
	\caption{Summary of the results of the simple dictionary-based approaches.}
	\centering{
		\begin{tabular}[h!]{|l|c|c|c|c|c|c|c|}
			\hline
			&&\multicolumn{2}{c|}{\bfseries Experiment 1} & \multicolumn{2}{c|}{\bfseries Experiment 2} & \multicolumn{2}{c|}{\bfseries Experiment 3}\\
			\hline
			\textbf{Sentiment} & \textbf{Actual Count} & \textbf{Count} & \textbf{\%}& \textbf{Count} & \textbf{\%}& \textbf{Count} & \textbf{\%} \\
			\hline
			Positive & 249 &69&28\%&123 & 49\% & 184&74\%\\
			Negative & 419 &180&43\%& 323 & 77\% &379 &90\%\\
			\hline
			Overall & 668&249&37\%& 446 & 67\% &563 &84\%\\
			\hline
		\end{tabular}
	}
	\label{tab:lexicon}
\end{table}

The results from the various experiments highlight the need for not only domain-specific sentiment prediction tools but also region-specific corpora.

The data set is named \textbf{\textit{LM-SA-2020}} representing {\it Loughran and McDonald Sentiment Word Lists} for \textit{South Africa}. 

A future improvement is to assess the sentiment for sentences where multiple sentiment-carrying words are present to evaluate the impact on sentiment prediction accuracy.

From the above results it appears that a simple dictionary based method to annotate the document headlines  prove more accurate than pre-trained sentiment analysers.

\subsubsection{Feature-based Approach: XGBoost} \label{xgb}

For model training, 80\% of the data set (of the 668 documents) were used. The accuracy of prediction was 81\% $\pm$4.4\%. The accuracy on the 20\% unseen data was also 81\%. Table~\ref{tab:xgb_summary} gives the recall and F1-score on the full data set and includes the results for the other approaches for comparison. The overall accuracy for the XGBoost model was 94\% using all headlines. 

\begin{table}[h!]
	\caption{Summary of the performance of the various annotation methods.}
	\centering{
	\small\addtolength{\tabcolsep}{5pt}
		\begin{tabular}[h]{|l|c|c|c|c|c|}
			\hline
			&&\multicolumn{2}{c|}{\bfseries Recall}&\multicolumn{2}{c|}{\bfseries F1-score}\\
			\hline
			& \textbf{Accuracy}  & \textbf{Pos}  & \textbf{Neg}  & \textbf{Pos}  & \textbf{Neg} \\
			\hline
			Lexicon & 84\% & 74\%&90\% &80\%&89\%\\
			TextBlob & 19\% & 24\% & 16\%&34\%&27\%\\
			Vader & 51\% & 40\% & 58\% &48\%&68\%\\
			XGBoost & 94\% & 86\% & 98\% & 91\% & 95\% \\
			\hline
			
		\end{tabular}
	}
	\label{tab:xgb_summary}
\end{table}



\begin{figure*}[h!]
	\centering
	\includegraphics[width=5in]{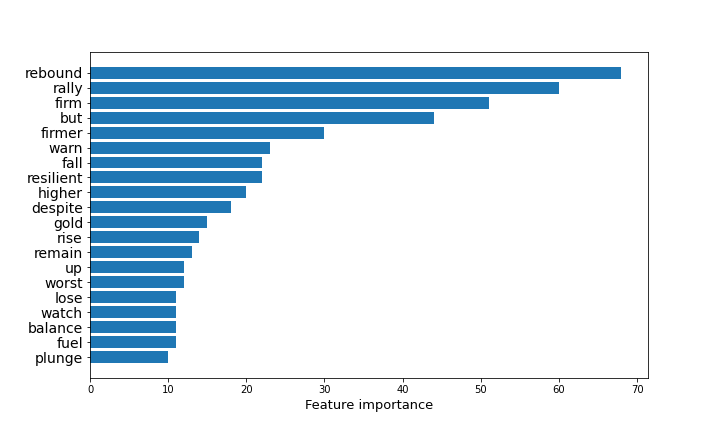}
	\caption{TF-IDF values as a function of occurrence.}
	\label{fig:XGBoost}
\end{figure*}

Figure~\ref{fig:XGBoost} shows the top 20 most important features of the XGBoost classifier. The words flagged are interpretable and useful.

The results given in this section lead to the conclusion that an XGBoost classifier as a simple traditional model performs very well for the given task and is recommended to be used as the sentiment prediction model.

For a potential future improvement, an approach that takes into account the sequence of words in a sentence should be evaluated for e.g. a recurrent neural network (RNN) such as a Long Short Term Memory (LSTM) with attention.

\subsection{Sentiment Correlation with Financial Performance}

\begin{figure*}[h!]
	\centering
	\hspace*{-1.3cm}
	\includegraphics[width=5.5in]{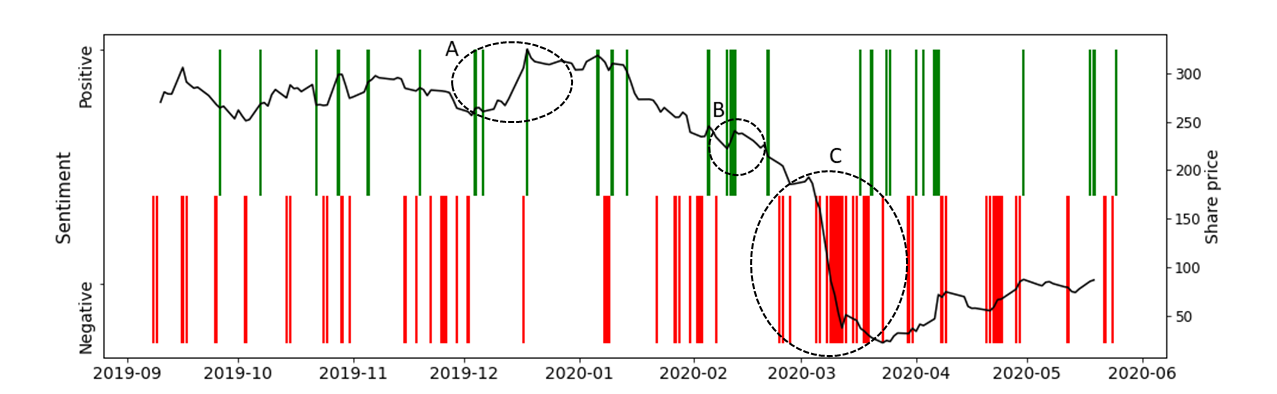}
	\caption{Sentiment prediction vs. share price for September 2019 - May 2020 using the XGBoost binary classifier.}
	\label{fig:corr}
\end{figure*}

In order to observe whether there is a noticeable trend between sentiment and share price, a time frame of the most recent six months was used. Figure~\ref{fig:corr} shows this trend. Periods A and B are periods where sentiment improved and was reflected by share price. Similarly Period C stands out through a significant amount of negative sentiments and a severe drop in share price.

The above results show promise that there are indeed periods where sentiment (from financial articles/documents) and share price correlate well.

An additional factor for consideration is the impact of lag when using share price
movement and it is recommended to be evaluated in future work.

It is also recommended to expand the sentiment prediction to include additional topics
and observe the correlation with share price. An alternative is to
extract topics according to the seven key drivers that impact reputation (Section \ref{intro}) and
apply weightings to an overall reputation score. 
Lastly, it is recommended to explore using either more fine-grained categories or
a continuous scale for sentiment.

\section{Model generalisation}
In order to understand how well the models generalise, it was required to use the developed model pipeline on unseen data from a different organisation. For this phase, data for the
corporate organisation, \textbf{\textit{Anglo American}}, was used. The aim was to determine whether language use in financial articles (mostly online news) follow the same pattern for different organisations. This will inform whether such models can be implemented on a larger scale or whether it is company-specific. 

As per the pipeline (Section \ref{method}), the following steps to predict sentiment (on headline or the first portion of a document) as well as to understand whether sentiment correlates with financial performance, were performed:
\begin{enumerate}
	\item Data collection and cleanup/pre-processing (a total of 1758 articles for the period June 2018 - May 2020)
	\item Filtering of data for financial documents with topic modelling (a total of 151 articles)
	\item Sentiment prediction using document titles:
	\begin{itemize}
		\item Using the updated dictionary to identify sentiment-carrying keywords (Section \ref{lexicon})
		\item Using the previously developed binary XGBoost classifier based on \textbf{\textit{Sasol}} data (without retraining) (Section \ref{xgb})
	\end{itemize}
	\item Graphically represent daily aggregated sentiments and share price
\end{enumerate}

Table~\ref{tab:comp_gen} gives the predicted sentiments using the dictionary-based and the XGBoost models. 

\begin{table}
	\caption{Comparison of sentiment predictions.}
	\centering{
		\begin{tabular}[t]{|l|c|c|c|c|}
			\hline
			& \multicolumn{2}{c|}{\bfseries Lexicon-based} & \multicolumn{2}{c|}{\bfseries XGBoost} \\
			\hline
			& Count&\%&Count&\%\\
			\hline
			\bfseries Positive & 49&32\% &43 &28\% \\
			\bfseries Negative & 74&49\% & 108 &72\%\\
			\bfseries Neutral & 20& 13\%& - &-\\
			\bfseries Other & 8& 5\% &-&-\\
			\hline
		\end{tabular}
	}
	\label{tab:comp_gen}
\end{table}

From Table~\ref{tab:comp_gen} it seems that the XGBoost classifier is more biased towards negative sentiments whereas the dictionary-based approach appears more balanced. There is only a 52\% agreement between the two models. Since there is no ground truth sentiment labels for the data, it was decided to manually evaluate the predicted sentiments to provide a more informed view. The following is an extract of the headline sentiment predictions using a binary XGBoost classifier as well as a dictionary-based approach. 

\begin{table}
	\caption{Comparison of sentiment predictions on \textbf{\textit{Anglo American}} data.}
	\centering{
		\begin{tabular}[t]{|l|c|c|}
			\hline
			\bfseries Sentence & \bfseries XGBoost & \bfseries Lexicon- \\
			& \bfseries classifier & \bfseries based \\
			\hline
			Aveng execs get R17.7m in bonuses - Moneyweb&	Negative&	Negative\\
			\hline
			Sharp (partial) recovery in share prices - Moneyweb	&Positive&	Positive\\
			\hline
			JSE tumbles as global growth fears spread $\mid$ Fin24&	Negative&	Negative\\
			\hline
			Anglo American replaces Deloitte with PwC as external &	Negative&	Litigious\\
			auditor after 20 years&&\\
			\hline
			Anglo Says S. Africa’s Eskom a Major Risk as It Mulls &	Negative&	Negative\\
			Growth - Bloomberg&&\\
			\hline
			Another Major Investor Leaves the Pebble Mine $\mid$ NRDC&	Negative&	Positive\\
			\hline
			Mining lobbies and the modern world: new issue of Mine 	&Positive&	Positive\\
			Magazine out now&&\\
			\hline
			BHP approach to Anglo CEO signals end of Mackenzie&Negative&	Positive\\
			era is nearing	&&\\
			\hline
			JSE tracks global markets higher on improved Wall &	Positive&	Positive\\
			Street data $\mid$ Fin24&&\\
			\hline
			Rand firms as dollar, stocks fall	&Negative&	Positive\\
			\hline
			Anglo American delivers 3.5-billion USD profit, declares &Negative&	Neutral\\
			final dividend	&&\\
			\hline
			Best Mining Stocks to Buy in 2020 $\mid$ The Motley Fool &Positive&	Positive\\
			\hline
			Rand firmer as dollar falls on rate cut bets	&Negative&	Positive\\
			\hline
			The new ministers in charge of the Amazon&	Positive&	Negative\\
			\hline
			Anglo American's Cutifani not thinking of retirement &Negative&	Negative\\
			as plots coup de grâce - Miningmx	&&\\
			\hline
			Pressure persists for resources stocks $\mid$ Fin24	&Positive&	Negative\\
			\hline
			Markets WRAP: Rand closes at R14.73/\$ $\mid$ Fin24	&Negative&	Neutral\\
			\hline
			See the top performers on the JSE in 2018 so far&	Positive&	Litigious\\
			\hline
			Anglo American seeks to avert revolt over chief's £14.6m &Negative&	Negative\\
			pay $\mid$ Business News $\mid$ Sky News	&&\\
			\hline
			Rand, stocks slip as investors await big Trump speech&	Negative&	Negative\\
			
			\hline
		\end{tabular}
	}
	\label{tab:samplesent_gen}
\end{table}

For a better understanding, TF-IDF vectors were determined for the \textbf{\textit{Anglo American}} document headlines and the vocabulary compared with the pre-trained vocabulary. Only 29\% of the words in this data set exists in the pre-trained vocabulary.





From the manual inspection it is concluded that the dictionary-based approach predict sentiments more accurately than the XGBoost classifier that was developed using \textbf{\textit{Sasol}} data. 

\begin{figure*}[h!]
	\centering
	\hspace*{-1.3cm}
	\includegraphics[width=5.6in]{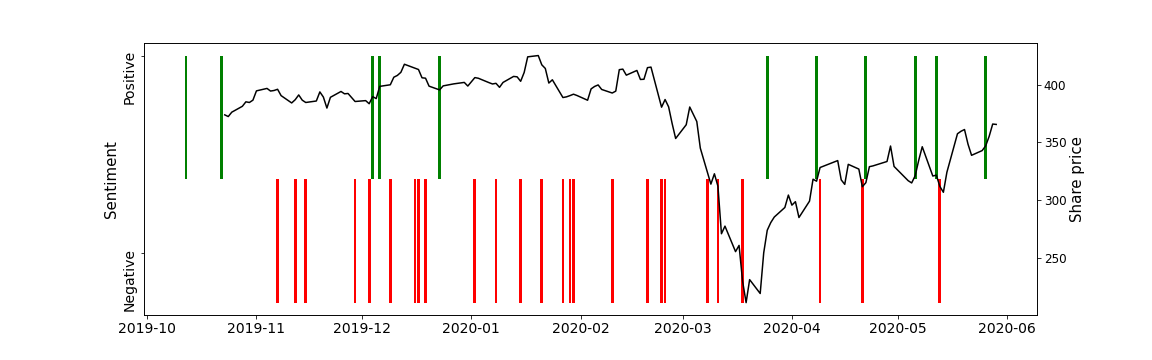}
	\caption{Sentiment prediction using a dictionary-based approach compared with Anglo American share price movement.}
	\label{fig:corr_dict}
\end{figure*}

Figure \ref{fig:corr_dict} shows the sentiment prediction and share price using a simple dictionary-based approach to identify sentiment-carrying words. It can be seen that there is an upward movement in share price corresponding to more positive sentiments (post April 2020).


Based on the above findings it is surmised that a XGBoost classifier trained on company-specific document titles may be too specific to extend to other industries. However, this can be improved by increasing the size of the data set to improve generalisation.

\section{Conclusions and Future Work}

Based on the findings it is concluded that natural language processing techniques can be used to predict the sentiment of financial articles in the South African context. Existing off-the-shelf sentiment analysers were evaluated and were found to underperform in predicting sentiment of South African finance-related articles with accuracies just above 50\%. Custom models using a simple lexicon-based approach or traditional machine learning such as a binary XGBoost classifier are well suited to the task and produced accuracies of 84\% and 94\% respectively. These models use document titles only.

Furthermore, an analysis showed there is a good correlation between predicted sentiments and financial performance (as represented by share price). The approach therefore shows promise, and with refinement, can be used to identify at risk periods for an organisation.

Lastly, the sentiment prediction model was evaluated using data from a different company to test how well it generalises. Since there were no ground truth data labels for this, a manual evaluation on a sample of the results was done. The dictionary-based approach the XGBoost classifier were compared and it was concluded that the former was better suited in this case. Sentiment predictions can be improved by increasing the size of the data set used in model development. Despite these shortcomings, a correlation between predicted sentiment and share price was still observed for certain periods. This substantiates the fact that the method has promise.\\

The main contributions made by this study are as follows: Developed an updated sentiment dictionary suitable for financial articles (the \textbf{\textit{LM-SA-2020}} data set and the accompanying data statement are publicly available) \cite{terblanche_marivate2021}.; Setting the foundation for expanding the work to include a broader sentiment prediction model that takes into account various topics and their contribution to overall sentiment as an indication of company reputation.; Progress towards an open-source library for financial sentiment analysis developed on South African data.

The following are some of the main recommendations for future work: A more sophisticated, streamlined process to update/expand the new data set - \textbf{\textit{LM-SA-2020}}.; Improve the model generalisation capability by increasing the size of the data set.; Investigate the impact of share price movement lag on the correlation with sentiment and enhance the understanding on whether there is a causal relationship between sentiment and financial performance.; Expand the sentiment prediction model to include additional topics (over and above financial documents).; Publish an open-source financial sentiment analysis tool that can be used on South African data.; Evaluate the performance of deep learning models - a very recent study indicated that transformers outperformed other sentiment analysis approaches and models in the domain of finance. It therefore warrants further investigation for possible application to this paper \cite{mishev2020evaluation}. 


%
%
%
\bibliographystyle{splncs04}
\bibliography{mit800}
%





\end{document}